\documentclass[journal]{IEEEtran}
\ifdefined\pdfshellescape\else
  \newcount\pdfshellescape
\fi
\usepackage{amsmath}
\usepackage{stfloats}
\usepackage[pdftex]{graphicx}
\usepackage{caption}
\usepackage{subfiles}
\usepackage{subfigure}
\usepackage{graphicx}

\usepackage{enumitem}
\usepackage{amsmath,amssymb,amsopn,amstext,amsfonts}
\usepackage{newtxtext,newtxmath}
\usepackage{cancel}
\usepackage[space]{cite}
\usepackage{balance}
\usepackage{color}
\usepackage[table]{xcolor}
\usepackage{algorithm}
\usepackage{algorithmic}
\usepackage{url}
\usepackage{booktabs}
\usepackage{threeparttable}
\usepackage[linkcolor=black,citecolor=black,urlcolor=black,colorlinks=true]{hyperref}
\usepackage{multirow}
\usepackage[table]{xcolor}
\usepackage{colortbl}
\usepackage{graphicx}
\usepackage{array}
\usepackage{verbatim}
\usepackage{makecell}
\usepackage{stackengine}
\usepackage{balance}
\usepackage{paracol}
\definecolor{mygreen}{RGB}{219, 234, 192}
\newcommand{\capyes}{\textcolor{green!60!black}{\ensuremath{\boldsymbol{\checkmark}}}}
\newcommand{\capno}{\textcolor{red}{\ensuremath{\boldsymbol{\times}}}}
\newcommand{\caphead}[2]{\shortstack[c]{\textbf{#1}\\[-0.15em]\textbf{#2}}}
\newcommand{\methodhead}{\textbf{Methods}}

\graphicspath{{./figures/}{./figure/}}

\DeclareGraphicsExtensions{.pdf,.png,.jpg,.PNG}
\title{\huge SCAN-Planner: Spatial Collision-Aware Local Planning for Route-Guided Long-Range Quadruped Navigation}
\author{

Han Zheng$^1$, Zhe Chen$^1$, Yiwen Fu$^1$, Ming Yang$^1$, and Tong Qin$^{1\ast}$
    \thanks{
        $^{1}$Shanghai Jiao Tong University, Shanghai, China.
        {\{hanzheng, qintong\}@sjtu.edu.cn}.
    }
    \thanks{
    {$^\ast$ is the Corresponding author. This work was supported by the Natural Science Foundation of Shanghai (Grant No. 24ZR1435600).}
    }
}

\begin{document}
\maketitle

\begin{abstract}
Quadruped robots are increasingly expected to navigate through narrow passages, cluttered indoor scenes, and large-scale 3D unstructured environments.
Existing local planners commonly approximate the robot using isotropic geometric inflation or rely on planar and elevation-map representations, leading to conservative motion in tight spaces and limited reasoning about overhanging structures.
This letter presents SCAN-Planner, a spatial collision-aware local planning framework for long-range quadruped navigation.
A yaw-aware twin-cylinder footprint is used to model the elongated robot body, enabling whole-body collision evaluation through sparse queries in an inflated 3D occupancy map.
We further introduce a projected A$^\ast$ search that generates collision-free guidance on an interpolated ground-following surface, with z-gradient suppression to avoid obstacles horizontally while maintaining vertical stability.
For large-scale deployment, a robot-centric sliding map with boundary fallback provides high-resolution local collision checking and recovery from local dead ends.
Simulation and real-world experiments demonstrate that SCAN-Planner generates safe, smooth, and efficient trajectories in dense clutter, 3D unstructured scenes, stair traversal, and long-range navigation tasks.
The code will be released at \href{https://github.com/wuyi2121/SCAN-Planner}{\textcolor{blue}{https://github.com/wuyi2121/SCAN-Planner}}.

\end{abstract}


\section{Introduction}
\label{sec:intro}
Quadruped robots are increasingly deployed for autonomous inspection, delivery, and service tasks in environments where wheeled robots are limited by discontinuous terrain and aerial robots are constrained by payload, endurance, or safety requirements.
Recent systems have demonstrated strong capabilities in field navigation~\cite{wellhausen2023artplanner}, multi-goal planning~\cite{chen2023smug}, and perceptive locomotion over rough terrain~\cite{grandia2023perceptive}, but they typically address terrain traversability, locomotion stability, or local obstacle avoidance as separate problems.
In practical long-range navigation, however, a quadruped robot must move through narrow passages, cluttered indoor scenes, staircases, and large-scale unstructured environments within a single planning framework.
This setting makes local planning particularly challenging: the planner must account for the yaw-dependent body footprint in tight spaces, reason about full 3D obstacles while preserving ground-following motion, and remain scalable when only a local map is available under coarse global route guidance.

\begin{figure}[!t]
    \centering
    \includegraphics[width=\columnwidth]{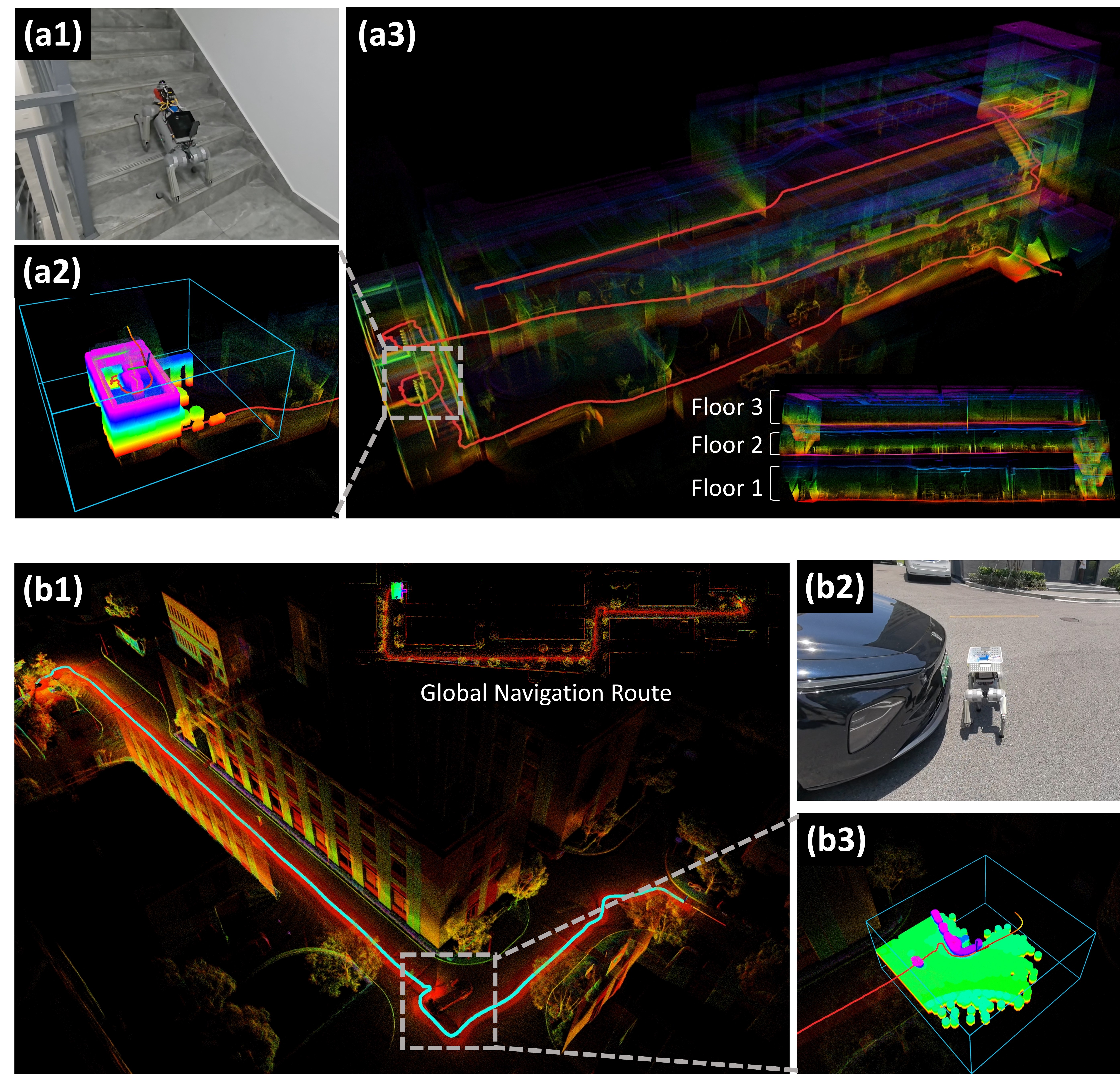}
    \caption{Real-world demonstrations of SCAN-Planner in challenging quadruped navigation tasks.
    (a1)--(a3) Multi-floor indoor navigation: the robot traverses staircases across different floors while performing local 3D collision checking, and the reconstructed point-cloud map shows the complete cross-floor trajectory. (b1)--(b3) Long-range outdoor navigation: the robot follows a coarse global navigation route in a large-scale campus environment, while SCAN-Planner maintains a robot-centric local occupancy map for real-time obstacle avoidance and route-guided trajectory generation.}
    \label{fig:real_teaser}
\end{figure}

The first difficulty lies in whole-body collision checking in narrow spaces.
Most real-time local planners achieve efficiency by modeling the robot as a point, sphere, or circle and inflating obstacles accordingly~\cite{zhou2020ego,zhou2019robust,tordesillas2022faster}.
Such a simplification is effective for compact robots, but it is inaccurate for quadruped bodies whose collision region changes with yaw.
A width-based inflation may be unsafe during turning, whereas a length-based inflation easily becomes over-conservative and blocks feasible narrow passages.
Whole-body planners have been developed to improve collision fidelity.
For example, RC-ESDF~\cite{geng2023robo} prebuilds a robot-centric signed distance field and jointly optimizes position and rotation, while 3D RoA-Planner~\cite{son2026roa} searches collision-free yaw ranges for confined-space quadruped planning.
Nevertheless, explicit yaw-state reasoning or rotation optimization increases planning complexity.
For onboard replanning, it is desirable to retain a compact position-trajectory optimization while evaluating collision safety with a yaw-aware body footprint.

The second difficulty comes from 3D unstructured scenes.
A 2D planner cannot distinguish a wall from a traversable step, and a 2.5D elevation map cannot faithfully represent overhanging tables, shelves, or multi-level structures around the robot body.
Recent ground-robot planners exploit richer 3D information, such as traversable planes~\cite{zhang2025traversableplanes} or multilevel terrain representations~\cite{li2025multilevel}.
However, directly treating a ground robot as a free-flying 3D body can introduce unnecessary vertical detours during local optimization.
For stable locomotion, obstacle avoidance should mainly deform the trajectory horizontally while preserving the vertical trend on slopes or stairs.

The third difficulty is long-range navigation with a local planner.
A fixed-horizon planner is efficient, but it is prone to local minima and dead ends when the target is far away and the map is only partially observed.
Large-scale autonomy systems therefore often rely on hierarchical guidance or coverage paths~\cite{cao2022aede,zhang2024falcon}, while robot-centric occupancy maps reduce memory cost by maintaining a local grid around the robot~\cite{ren2023rogmap}.
However, route guidance, sliding-map boundaries, high-resolution collision checking, and dead-end recovery must be handled consistently.
Otherwise, a local planner may fail when the subgoal lies behind a newly observed obstacle or outside the current map boundary.

To bridge these gaps, we propose SCAN-Planner, a spatial collision-aware local planner for route-guided long-range quadruped navigation.
SCAN-Planner keeps the efficiency of position-only B-spline optimization, while making initialization, collision segmentation, rebound search, safety checking, and execution consistent with a yaw-aware quadruped footprint.
It further combines a projected A$^\ast$ search with z-gradient suppression, a robot-centric sliding occupancy map, and global coarse-route guidance to support narrow-space, 3D, and long-range navigation.
The main contributions of this paper are summarized as follows:
\begin{itemize}[leftmargin=*]
    \item We propose a yaw-aware footprint representation for efficient whole-body collision checking of quadruped robots in narrow and 3D unstructured spaces.
    \item We introduce a projected A$^\ast$ planning scheme with z-gradient suppression, enabling horizontal obstacle avoidance while maintaining vertical stability in 3D planning.
    \item We develop a robot-centric sliding map that supports high-resolution local collision checking, scalable planning in large-scale environments, and robust dead-end recovery.
    \item We integrate local trajectory optimization with global coarse-route guidance, and validate the reliability and generalizability of the system through extensive real-world experiments.
\end{itemize}

\begin{table*}[t]
\caption{Capability Comparison for Recent Local Planners}
\label{tab:capability_comparison}
\centering
\small
\renewcommand{\arraystretch}{0.96}
\setlength{\tabcolsep}{2.0pt}
\begin{tabular}{@{}>{\raggedright\arraybackslash}m{0.17\textwidth}*{6}{>{\centering\arraybackslash}m{0.105\textwidth}}@{}}

\toprule
\methodhead &
\caphead{Terrain}{Traversal} &
\caphead{Overhang}{Obstacles} &
\caphead{Ground}{Following} &
\caphead{Smooth}{Trajectory} &
\caphead{Long-Range}{Navigation} &
\caphead{Yaw-Aware}{Body} \\
\midrule
EGO-Planner-2D~\cite{zhou2020ego} & \capno & \capno & \capyes & \capyes & \capno & \capno \\
EGO-Planner-3D~\cite{zhou2020ego} & \capyes & \capyes & \capno & \capyes & \capno & \capno \\
ART-Planner~\cite{wellhausen2023artplanner} & \capyes & \capno & \capyes & \capno & \capno & \capyes \\
CMU-Planner~\cite{cao2022aede} & \capyes & \capno & \capyes & \capno & \capno & \capno \\
SCAN-Planner (Ours) & \capyes & \capyes & \capyes & \capyes & \capyes & \capyes \\
\bottomrule
\end{tabular}
\end{table*}

\section{Related Work}
\label{sec:related}

\subsection{Local Trajectory Planning}

Local trajectory planning in cluttered environments is commonly addressed through kinodynamic search, corridor-constrained optimization, and gradient-based trajectory deformation. Kinodynamic planners generate dynamically feasible trajectories by searching in the state space~\cite{zhou2019robust,tordesillas2022faster}, while corridor-based methods optimize polynomial or MINCO trajectories within safe convex regions~\cite{wang2022geometrically,han2021fast}. Gradient-based replanning methods further improve efficiency by optimizing smoothness, feasibility, and collision costs. RAPTOR~\cite{zhou2021raptor} improves replanning quality with topological path guidance and perception-aware refinement, while EGO-Planner~\cite{zhou2020ego} avoids explicit ESDF construction by lazily generating rebound constraints from collision-free guiding paths. These methods provide efficient local trajectory generation in cluttered environments and have been widely adopted for real-time robotic navigation.

Despite their efficiency, most local planners rely on point-mass or isotropically inflated robot models for collision checking. Such abstractions are insufficient for elongated quadruped bodies whose footprint varies with yaw, especially in narrow spaces with lateral or overhanging obstacles. Whole-body collision-aware methods, such as RC-ESDF~\cite{geng2023robo}, improve collision fidelity by reasoning over robot-centric distance fields and orientation states, but the additional state dimensions increase onboard replanning complexity. Efficient quadruped local planning therefore requires a collision model that is both body-aware and lightweight for real-time optimization in 3D cluttered environments.

\subsection{Quadruped Robot Navigation}

Quadruped navigation requires terrain perception, locomotion feasibility, and navigation-level planning. Existing legged navigation methods estimate traversable regions using reachability reasoning, template learning, or elevation mapping~\cite{wellhausen2021rough,miki2022elevation}. Other works learn policy-specific traversability from volumetric occupancy~\cite{frey2022locomotion}, improve terrain-adaptive control with nonlinear MPC~\cite{grandia2023perceptive}, or train learned policies for challenging terrain traversal~\cite{dong2025marg}. These methods provide strong locomotion and traversability capabilities, but they mainly focus on local terrain feasibility rather than smooth trajectory planning in cluttered 3D spaces. As a result, their outputs are not directly formulated as continuous, collision-aware local trajectories for navigation.

Navigation-oriented systems further incorporate terrain and body constraints into higher-level planning. ART-Planner~\cite{wellhausen2023artplanner} combines traversability-informed sampling with MPC for robust field navigation, while SMUG Planner~\cite{chen2023smug} performs safe multi-goal planning with robot-specific validity checking. Recent methods also exploit foot-end terrain features~\cite{li2026stable} or search collision-free yaw ranges for confined-space maneuvering~\cite{son2026roa}.
However, deploying quadruped navigation in large-scale environments remains challenging under bounded-memory local mapping. The finite map window may truncate obstacles and route information at the map boundary, while long-range route following inevitably encounters dead-end regions where local planners can become trapped in local minima.

\section{System Overview}
\label{sec:overview}
\begin{figure*}[t]
    \centering
    \includegraphics[width=0.90\textwidth]{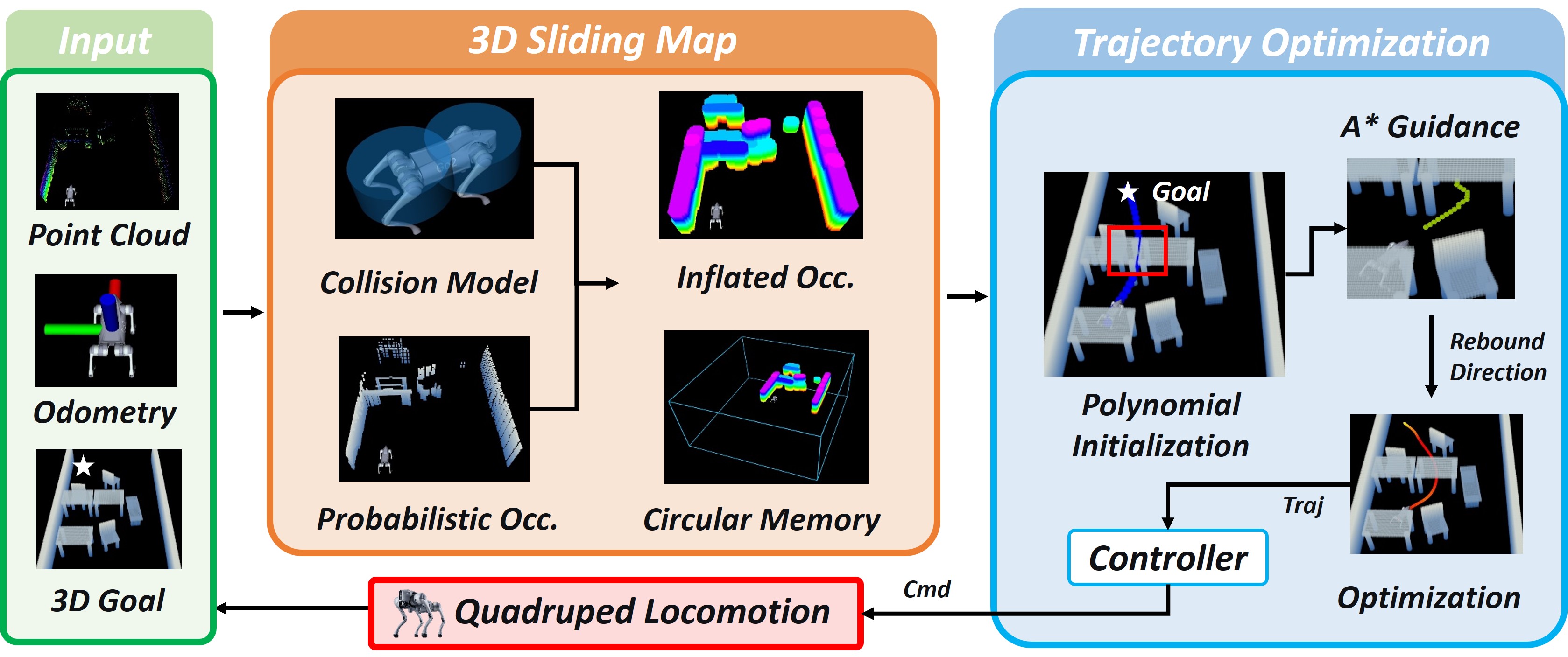}
    \caption{System architecture of SCAN-Planner.
The system takes point clouds, odometry, and a route-guided 3D local goal as inputs. A robot-centric 3D sliding map maintains probabilistic occupancy, yaw-aware inflated occupancy, and circular-buffer memory for high-resolution local collision checking. Based on the map representation, the trajectory optimization module generates a height-regularized polynomial initialization, performs projected A$^\ast$ guidance and rebound-direction construction when collisions are detected, and optimizes a smooth collision-free trajectory for execution by the quadruped locomotion controller.}
    \label{fig:framework}
\end{figure*}

The proposed system is shown in Fig.~\ref{fig:framework}.
SCAN-Planner takes the point cloud, odometry, and a 3D goal from the global route as inputs.
A robot-centric 3D sliding map maintains probabilistic occupancy, inflated occupancy, and circular-buffer memory for high-resolution local collision checking and large-scene operation.
The trajectory planning module first generates a height-regularized polynomial initialization, detects unsafe segments using the yaw-aware footprint, searches for projected A$^\ast$ guidance when needed, and optimizes a smooth B-spline trajectory.
The optimized trajectory is sent to the controller and executed by the quadruped locomotion module.

\section{Yaw-Aware Local Planning}
\label{sec:local_planning}

\subsection{Trajectory Representation}
\label{subsec:trajectory_representation}

For a quadruped robot whose heading can be decoupled from its direction of motion,
we optimize only a position trajectory \(\mathbf{p}(t)\in\mathbb{R}^3\), and
assign an induced yaw angle \(\psi(t)\in\mathbb{S}^1\) for footprint collision
checking and execution. Specifically, the position trajectory is parameterized by a uniform B-spline with degree \(p_{\mathrm b}\), knot span \(\Delta t\), and \(N_{\mathrm c}\) control points \(\{\mathbf{Q}_i\}_{i=0}^{N_{\mathrm c}-1}\), which is a piecewise polynomial uniquely determined by its control points.
In practice, we use a cubic B-spline with \(p_{\mathrm b}=3\), whose duration is \(T=(N_{\mathrm c}-p_{\mathrm b})\Delta t\).
The control points of the velocity, acceleration, and jerk curves are obtained by:
\begin{equation}
\mathbf{V}_i = \frac{\mathbf{Q}_{i+1}-\mathbf{Q}_i}{\Delta t},\;
\mathbf{A}_i = \frac{\mathbf{V}_{i+1}-\mathbf{V}_i}{\Delta t},\;
\mathbf{J}_i = \frac{\mathbf{A}_{i+1}-\mathbf{A}_i}{\Delta t}.
\label{eq:derivatives}
\end{equation}

We optimize a subset of \(N_{\mathrm c}-2p_{\mathrm b}\) interior control points,
\(\{\mathbf{Q}_{p_{\mathrm b}},\mathbf{Q}_{p_{\mathrm b}+1},\ldots,\mathbf{Q}_{N_{\mathrm c}-p_{\mathrm b}-1}\}\).
The first and last \(p_{\mathrm b}\) control points are kept fixed because they determine the boundary state.

For a quadruped robot, aligning the body heading with the local direction of motion reduces the lateral footprint in narrow passages. Therefore, instead of optimizing yaw independently, we induce the yaw used for whole-body collision checking from the local tangent of the position trajectory:
\begin{equation}
\psi_i =
\operatorname{atan2}
\left(
\mathbf{e}_{\mathrm y}^{\top}(\mathbf{Q}_{i+1}-\mathbf{Q}_{i-1}),
\mathbf{e}_{\mathrm x}^{\top}(\mathbf{Q}_{i+1}-\mathbf{Q}_{i-1})
\right).
\label{eq:induced_yaw}
\end{equation}
where \(\mathbf{e}_{\mathrm x}=[1,0,0]^{\top}\) and \(\mathbf{e}_{\mathrm y}=[0,1,0]^{\top}\).
Here, \(\mathbf{Q}_{i+1}-\mathbf{Q}_{i-1}\) provides a finite-difference
approximation of the local motion direction associated with the \(i\)-th
checking point, as shown in Fig.~\ref{fig:inflation}. This coupling avoids
introducing additional yaw control variables while keeping the footprint
collision check consistent with tangent-following quadruped execution.

\subsection{Whole-Body Collision Check}
\label{subsec:wb_collision}
\begin{figure}[t]
    \centering
    \includegraphics[width=\linewidth]{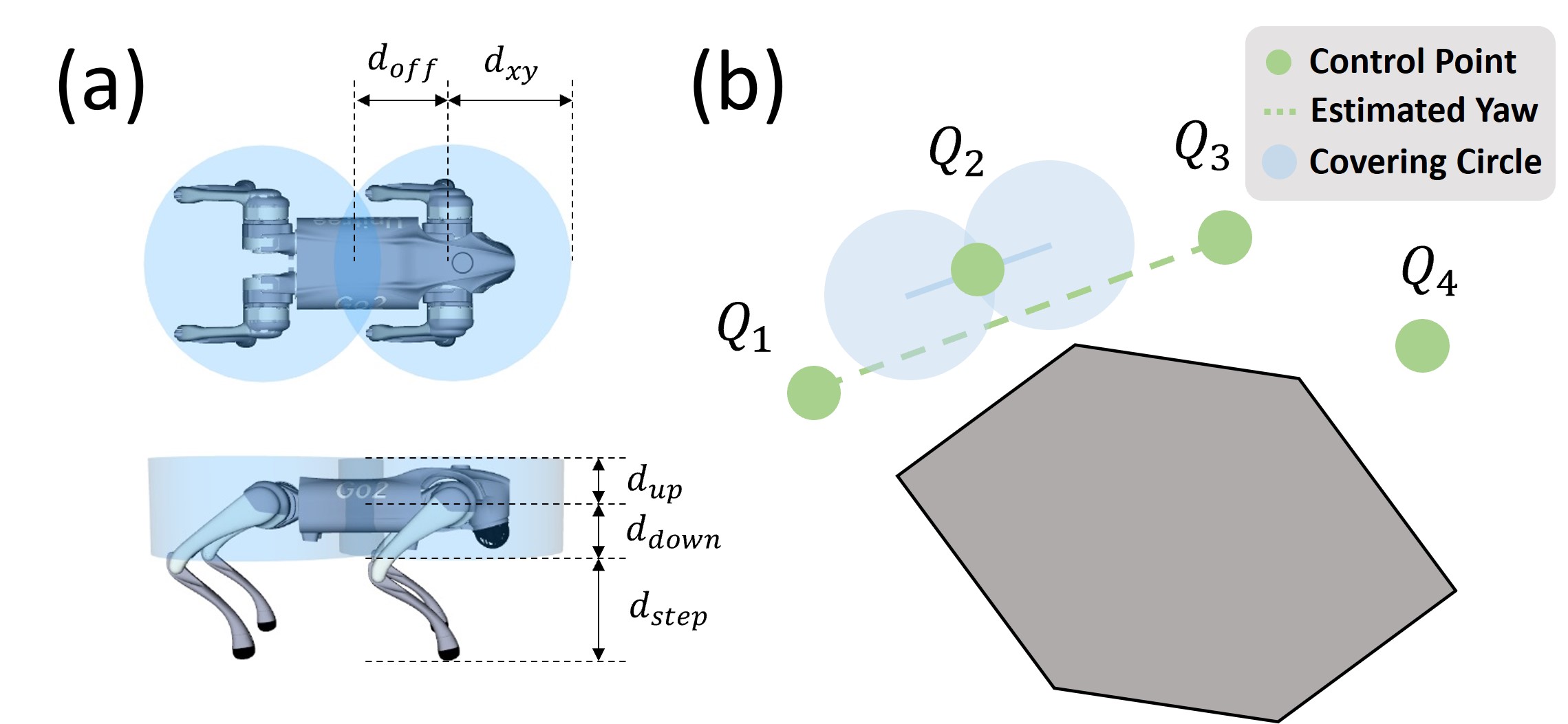}
    \caption{Yaw-aware footprint for whole-body collision checking. (a) The quadruped body is covered by two vertical cylindrical primitives with longitudinal offset and clearance parameters. (b) Along the B-spline control points, collision checking is reduced to querying the transformed cylinder centers in the inflated occupancy map.}
    \label{fig:inflation}
\end{figure}

To improve efficiency, we approximate the elongated robot body by two
vertical cylinders fixed in the body frame. After inflating the occupancy map
by the cylinder radius and vertical clearances, yaw-aware whole-body collision
checking reduces to finite-point queries of the transformed cylinder centers.
As shown in Fig.~\ref{fig:inflation}, the two cylinder centers are placed along
the body longitudinal axis:
\begin{equation}
\mathbf{s}_{\mathrm b,1}=[d_{\mathrm{off}},0,0]^{\top},\qquad
\mathbf{s}_{\mathrm b,2}=[-d_{\mathrm{off}},0,0]^{\top},
\label{eq:body_centers}
\end{equation}
where \(d_{\mathrm{off}}\) is the longitudinal offset from the robot center.
Each cylinder is parameterized by a horizontal clearance \(d_{\mathrm{xy}}\), an upward clearance \(d_{\mathrm{up}}\), and a downward clearance \(d_{\mathrm{down}}\); \(d_{\mathrm{step}}\) denotes the maximum traversable step height used to set the lower clearance.
The upward clearance \(d_{\mathrm{up}}\) is selected to cover the maximum vertical extent of the robot body and mounted hardware with a small safety margin, enabling the planner to distinguish truly blocking overhead obstacles from structures with sufficient clearance.
The downward clearance \(d_{\mathrm{down}}\) is set to the nominal robot-center height minus \(d_{\mathrm{step}}\), thereby excluding traversable low obstacles from the inflated region while still rejecting obstacles that exceed the robot's terrain-crossing capability.

Given a checking control point \(\mathbf{Q}_k\) and its induced yaw \(\psi_k\), the cylinder centers in the world frame are
\begin{equation}
\mathbf{s}_{k,j}
=
\mathbf{R}_z(\psi_k)\mathbf{s}_{\mathrm b,j}+\mathbf{Q}_k,
\quad j\in\{1,2\}.
\label{eq:world_center}
\end{equation}
We maintain a 3D occupancy map inflated according to the above cylindrical primitive.
Thus, whole-body collision checking only requires querying the two transformed cylinder centers:
\begin{equation}
H_k(\mathbf{Q}_k,\psi_k)
=
\max_{j\in\{1,2\}}
\chi(\mathbf{s}_{k,j}),
\label{eq:body_occ}
\end{equation}
where \(\chi(\cdot)\) denotes the inflated occupancy value.
The robot is considered collision-free at the checking control point \(\mathbf{Q}_k\) if \(H_k(\mathbf{Q}_k,\psi_k)=0\).

\subsection{Objective Functions}
\label{subsec:objective}

The optimization problem is formulated as follows:
\begin{equation}
\min_{\mathbf{Q}}
J
=
\lambda_{\mathrm c} J_{\mathrm c}
+
\lambda_{\mathrm s} J_{\mathrm s}
+
\lambda_{\mathrm f} J_{\mathrm f},
\label{eq:objective}
\end{equation}
where \(J_{\mathrm c}\) is the rebound collision penalty, and \(J_{\mathrm s}\) and \(J_{\mathrm f}\) denote
the smoothness and feasibility penalties, respectively. The coefficients
\(\lambda_{\mathrm c}\), \(\lambda_{\mathrm s}\), and \(\lambda_{\mathrm f}\) are the corresponding weights.

\subsubsection{Height-Regularized Initialization}
\begin{figure}[t]
    \centering
    \includegraphics[width=0.90\linewidth]{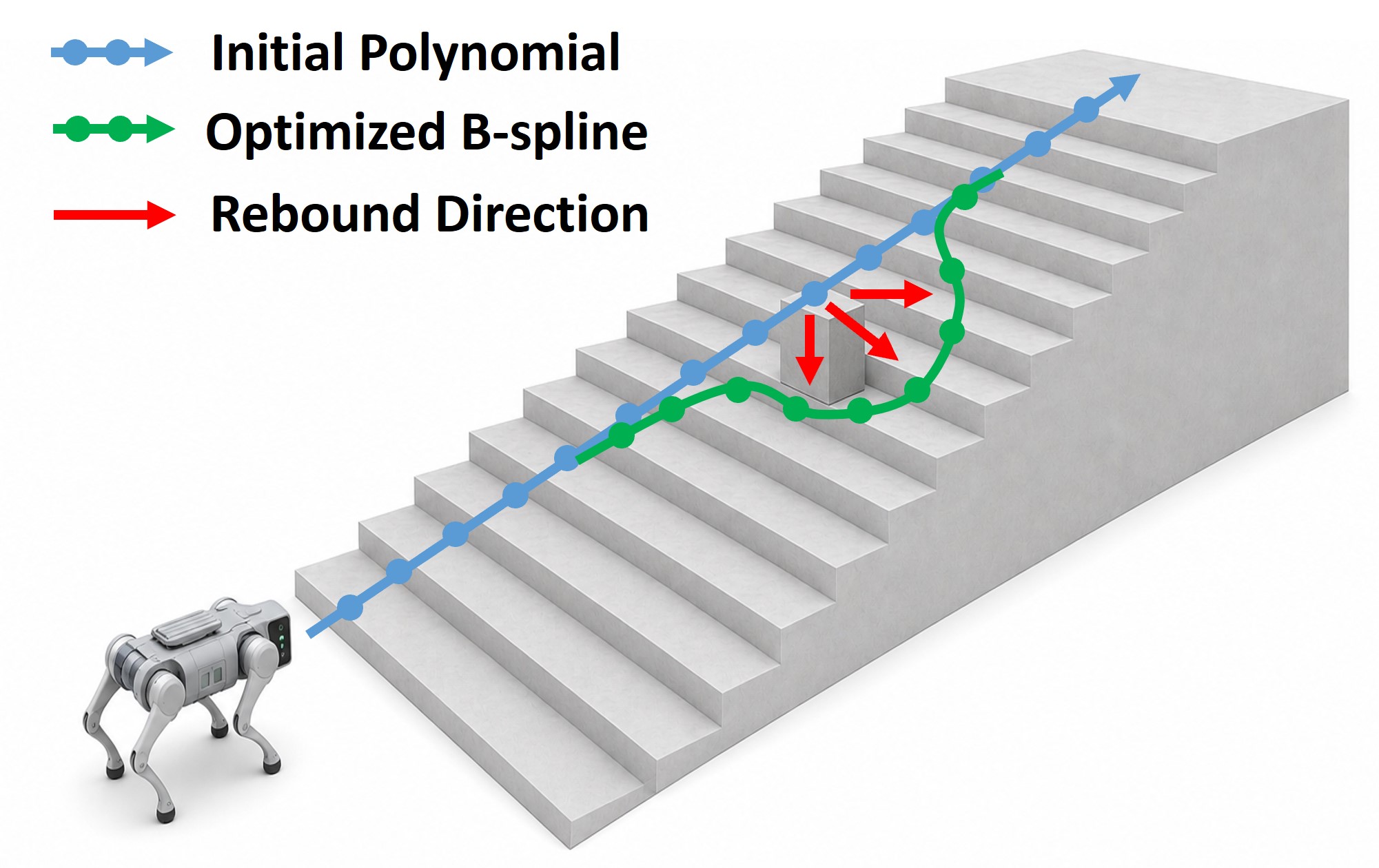}
    \caption{Projected rebound deformation with z-gradient suppression on stair-like terrain. The height-regularized polynomial initialization (blue) provides the nominal vertical profile, while the optimized B-spline (green) avoids the obstacle through horizontal deformation guided by rebound gradients (red), reducing unnecessary vertical oscillation.}
    \label{fig:gradient}
\end{figure}

The initial path is generated from a polynomial trajectory.
To avoid unnecessary vertical oscillation in 3D planning, we regularize the height of the sampled path before B-spline parameterization.
Let \(\mathcal{R}=\{\mathbf{r}_0,\mathbf{r}_1,\ldots,\mathbf{r}_{N_{\mathrm p}}\}\) be the raw samples of the polynomial path.
We first compute the accumulated horizontal arc length:
\begin{equation}
\ell_i =
\sum_{m=1}^{i}
\left\|
\mathbf{r}_{m,\mathrm{xy}}
-
\mathbf{r}_{m-1,\mathrm{xy}}
\right\|,
\quad L=\ell_{N_{\mathrm p}} .
\label{eq:xy_length}
\end{equation}
Then the height assigned to the \(i\)-th sample is
\begin{equation}
z_i = z_{\mathrm s}+\alpha_i(z_{\mathrm g}-z_{\mathrm s}),
\quad
\alpha_i =
\begin{cases}
\ell_i/L, & L>0,\\
i/N_{\mathrm p}, & L=0,
\end{cases}
\label{eq:z_reference}
\end{equation}
where \(z_{\mathrm s}\) and \(z_{\mathrm g}\) are the start and local target heights.
The height-regularized path preserves the \(xy\)-projection of the initial path,
while its \(z\)-coordinate follows a linear profile along the accumulated
horizontal arc length. These height-adjusted samples are used to parameterize the B-spline,
providing a stable vertical profile before rebound optimization. As shown in
Fig.~\ref{fig:gradient}, this encourages obstacle avoidance mainly through
lateral deformation while suppressing unnecessary vertical oscillation.

\subsubsection{Collision Penalty}

The collision penalty is constructed only for colliding segments detected by the yaw-aware footprint check in Eq.~\eqref{eq:body_occ}.
For each colliding
segment, a collision-free path \(\Gamma\) is generated by the projected A$^\ast$
search. Following the rebound formulation \cite{zhou2020ego}, each affected control point
\(\mathbf{Q}_i\) is assigned one or more \(\{\mathbf{a},\mathbf{v}\}\) pairs,
where \(\mathbf{a}_{ij}\) is an anchor point and \(\mathbf{v}_{ij}\) is the
corresponding unit rebound direction.
The obstacle distance from
\(\mathbf{Q}_i\) to the \(j\)-th obstacle is defined as
\begin{equation}
d_{ij}
=
(\mathbf{Q}_i-\mathbf{a}_{ij})^{\top}\mathbf{v}_{ij}.
\label{eq:obstacle_distance}
\end{equation}
The collision penalty is then written as
\begin{equation}
J_{\mathrm c}
=
\sum_{i,j}
\rho(c_{ij}),
\label{eq:collision_cost}
\end{equation}
where \(c_{ij}=s_{\mathrm f}-d_{ij}\), \(s_{\mathrm f}\) is the safety clearance, and
\(\rho(\cdot)\) is a twice continuously differentiable one-sided penalty. The corresponding gradient is
\begin{equation}
\frac{\partial J_{\mathrm c}}{\partial \mathbf{Q}_i}
=
-
\sum_j
\rho'(c_{ij})\mathbf{v}_{ij}.
\label{eq:collision_gradient}
\end{equation}
The binary occupancy indicator in Eq.~\eqref{eq:body_occ} is not differentiated; it is used only
to detect colliding segments and lazily construct the \(\{\mathbf{a},\mathbf{v}\}\)
pairs needed for the smooth collision penalty.

\subsubsection{Smoothness and Feasibility Penalty}
The smoothness penalty minimizes the squared norm of jerk control points:
\begin{equation}
J_{\mathrm s}
=
\sum_k
\|\mathbf{J}_k\|^2 .
\label{eq:smoothness_cost}
\end{equation}
To limit the velocity and acceleration along the trajectory, the feasibility penalty uses a one-sided squared penalty and is written as:
\begin{equation}
J_{\mathrm f}
=
\sum_k
\mathcal{F}(\|\mathbf{V}_k\|,v_{\mathrm m})
+
\sum_k
\mathcal{F}(\|\mathbf{A}_k\|,a_{\mathrm m}),
\label{eq:feasibility_cost}
\end{equation}
where \(v_{\mathrm m}\) and \(a_{\mathrm m}\) are the maximum velocity and acceleration, respectively.
The function \(\mathcal{F}(c,c_{\mathrm m})=\max(c-c_{\mathrm m},0)^2\) is a one-sided squared penalty.
In implementation, the vertical component of the optimization update is suppressed, encouraging obstacle avoidance mainly through horizontal deformation.
Since yaw is induced from the position trajectory, no yaw control points are
optimized. During execution, the physical heading is regulated to track the
planned tangent direction.

\section{3D Sliding Map}
\label{sec:sliding_map}

\subsection{Sliding Strategy}
\label{subsec:sliding_strategy}
The local map is maintained as a uniform 3D occupancy grid. Following common occupancy-mapping practice~\cite{ren2023rogmap,tang2026memory}, raycasting updates the log-odds occupancy value of each voxel with bounded probability clamping.
The upper clamp \(p_{\max}\) prevents occupied voxels from becoming
over-confident, while the lower clamp \(p_{\min}\) allows free or unknown
regions to be updated rapidly when new observations arrive. This update mechanism
supports both persistent structures and transient obstacles in complex environments.

\begin{figure}[t]
    \centering
    \includegraphics[width=0.90\linewidth]{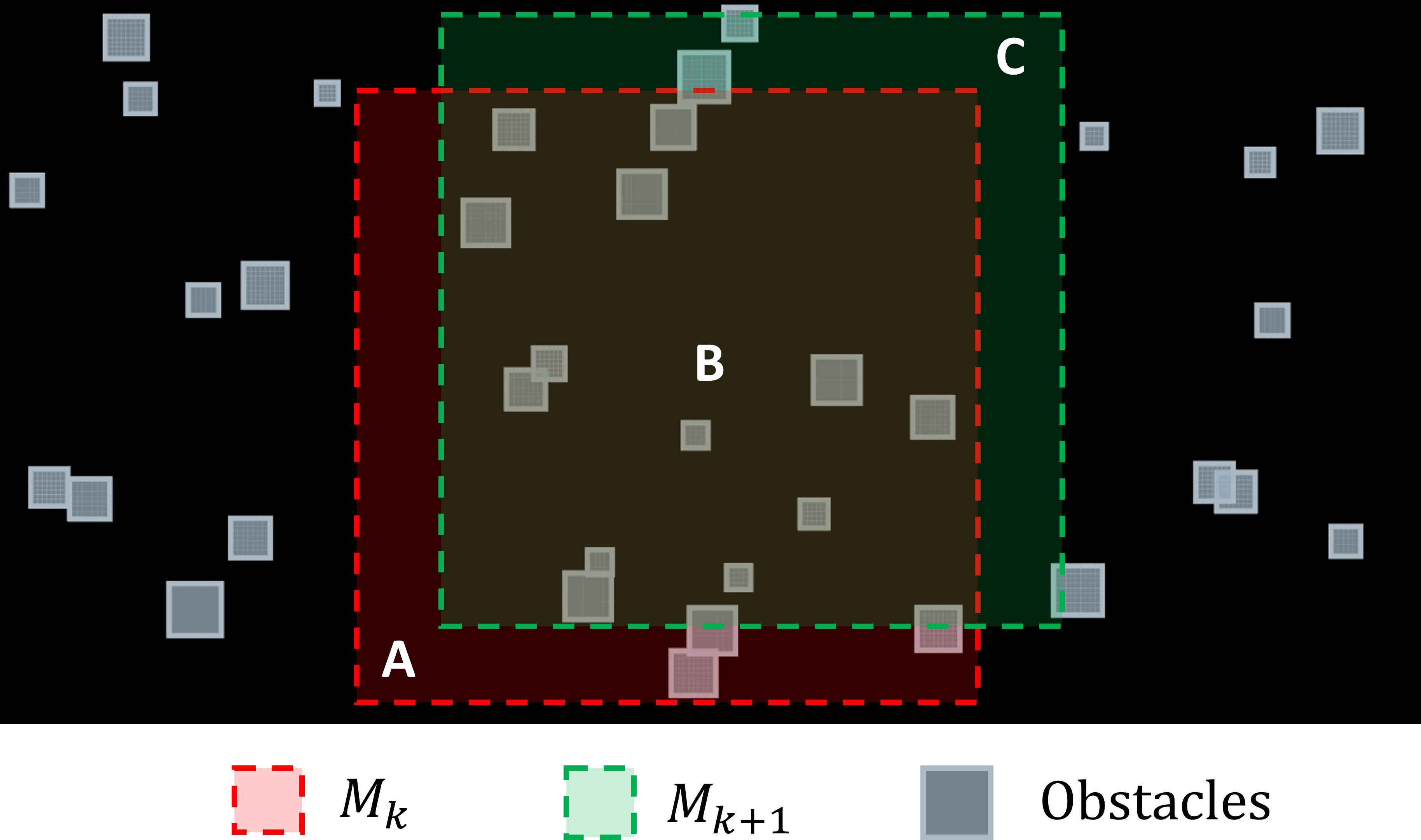}
    \caption{Sliding map update. When the active map shifts from \(\mathcal{M}_k\) (red) to \(\mathcal{M}_{k+1}\) (green), the overlapped region \(B\) is maintained without memory relocation, the outgoing region \(A\) is released, and the incoming region \(C\) is reset for newly observed voxels.}
    \label{fig:sliding_map}
\end{figure}

The map is implemented as a fixed-size sliding window centered around the
robot.
As illustrated in Fig.~\ref{fig:sliding_map}, when the sliding map shifts from \(\mathcal{M}_k\) to \(\mathcal{M}_{k+1}\), the overlapping region \(B\) is preserved, while
the outgoing region \(A\) releases circular-buffer addresses that are reused by
the incoming region \(C\).
Only the recycled addresses are reset and
reinitialized, and the buffer contents associated with the overlapping region are
kept unchanged.
Following ROG-Map~\cite{ren2023rogmap}, the inflated occupancy layers are updated incrementally when
voxel occupancy states change rather than
rebuilding the whole inflated map. This zero-copy sliding mechanism keeps memory
bounded and enables high-resolution local mapping during long-range navigation.

\subsection{Projected A$^\ast$ Guidance}
\label{subsec:planar_search}

When a colliding segment is detected, a collision-free path \(\Gamma\) is
searched between the entry point \(\mathbf{x}^{\mathrm{in}}\) and the exit point
\(\mathbf{x}^{\mathrm{out}}\) to construct the \(\{\mathbf{a},\mathbf{v}\}\) pairs.
Instead of searching the full 3D grid, we restrict A$^\ast$ expansion to an
inclined search surface whose height is interpolated from the two endpoints.
For a candidate horizontal grid point \(\mathbf{x}_{\mathrm{xy}}\), the interpolation
ratio is
\begin{equation}
\beta(\mathbf{x}_{\mathrm{xy}})
=
\Pi_{[0,1]}
\left(
\frac{
(\mathbf{x}_{\mathrm{xy}}-\mathbf{x}^{\mathrm{in}}_{\mathrm{xy}})^{\top}
(\mathbf{x}^{\mathrm{out}}_{\mathrm{xy}}-\mathbf{x}^{\mathrm{in}}_{\mathrm{xy}})
}{
\|\mathbf{x}^{\mathrm{out}}_{\mathrm{xy}}-\mathbf{x}^{\mathrm{in}}_{\mathrm{xy}}\|^2
}
\right),
\label{eq:search_ratio}
\end{equation}
where \(\Pi_{[0,1]}(\eta)=\min\{1,\max\{0,\eta\}\}\) denotes projection onto the interval \([0,1]\).
The corresponding height is
\begin{equation}
z(\mathbf{x}_{\mathrm{xy}}) =
z^{\mathrm{in}}+\beta(\mathbf{x}_{\mathrm{xy}})(z^{\mathrm{out}}-z^{\mathrm{in}}).
\label{eq:search_height}
\end{equation}
Thus, each search node is represented by its horizontal grid coordinate and the
interpolated height.

During neighbor expansion, we set the checking yaw to
\(\psi_{\Delta}=\operatorname{atan2}(\Delta y,\Delta x)\), and accept a
candidate node only when the yaw-aware collision indicator in
Eq.~\eqref{eq:body_occ} is zero. This projected search suppresses unnecessary
vertical detours and keeps the rebound guidance consistent with ground-robot
motion.

\begin{figure}[t]
    \centering
    \includegraphics[width=\linewidth]{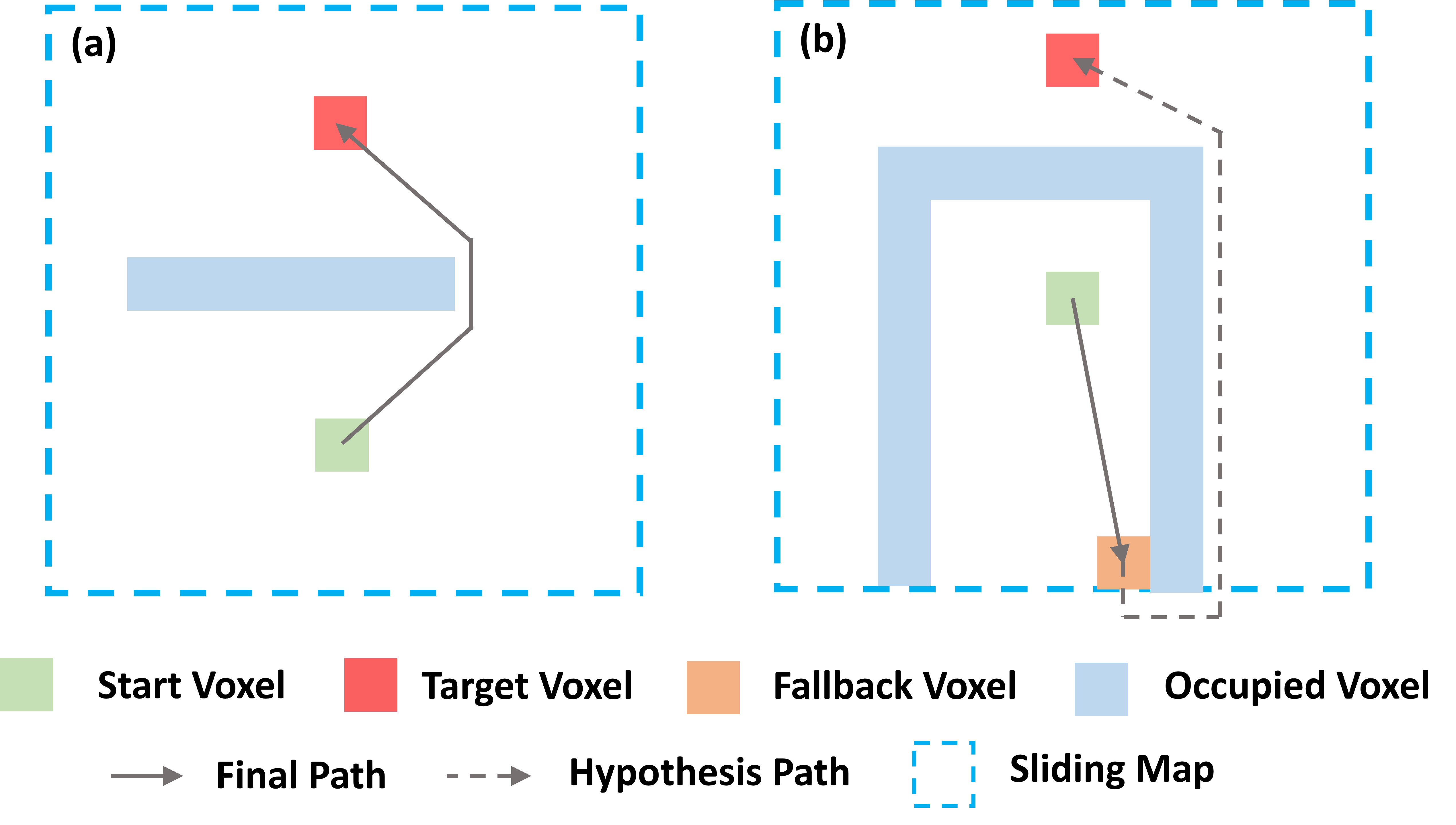}
    \caption{Projected A$^\ast$ guidance and boundary fallback. (a) When the target is reachable inside the sliding map, the final path detours around occupied voxels. (b) In a dead-end case, a hypothesis path is allowed to enter the virtual free layer outside the map boundary, and the exit boundary voxel is selected as a fallback target for a bounded recovery motion.}
    \label{fig:astar_fallback}
\end{figure}

\subsection{Dead-End Recovery}
To improve robustness in dead-end cases, we further adopt a boundary fallback strategy, as illustrated in Fig.~\ref{fig:astar_fallback}(b).
During the rebound search, the sliding map is temporarily augmented with a one-voxel-wide virtual free layer outside its boundary. This layer is used only for hypothesis-path generation and is not regarded as executable free space.
If the original target cannot be reached inside the sliding map, the projected A$^\ast$ search is allowed to enter the virtual layer, and the boundary voxel where the hypothesis path leaves the sliding map is selected as a fallback voxel.
The local target is then replaced by this fallback voxel, and the final path is generated within the real sliding map using the yaw-aware twin-cylinder collision check.
This mechanism converts a dead-end planning failure into a bounded recovery motion toward the map boundary.

\begin{figure*}[!t]
    \centering
    \includegraphics[width=0.90\textwidth]{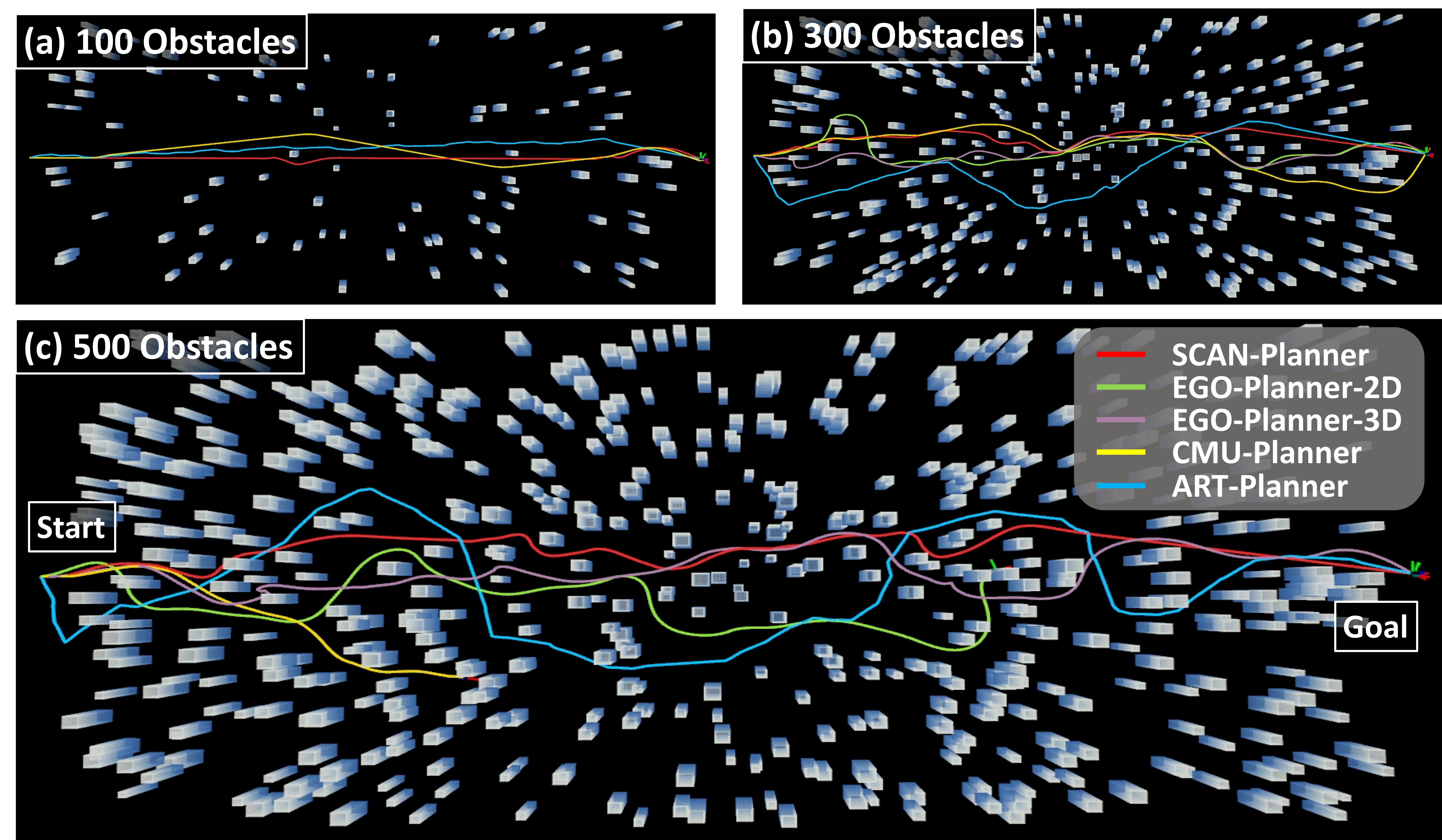}
    \caption{Simulation comparison in dense random obstacle fields. The planners are evaluated in environments with increasing obstacle density: (a) 100 obstacles, (b) 300 obstacles, and (c) 500 obstacles. The trajectories of SCAN-Planner, EGO-Planner-2D, EGO-Planner-3D, CMU-Planner, and ART-Planner are shown from start to goal. As the clutter density increases, SCAN-Planner generates a smooth and direct collision-free trajectory while maintaining yaw-aware body clearance, whereas the baseline methods exhibit larger detours, unstable route selection, or reduced efficiency in narrow passages.}
    \label{fig:compare_2d}
\end{figure*}

\section{Experimental Results}
\label{sec:exp}

\subsection{Experimental Setup}
\label{subsec:setups}

We evaluate SCAN-Planner against four representative local planning baselines
using the MARSIM simulator~\cite{kong2022marsim}: EGO-Planner-2D and EGO-Planner-3D~\cite{zhou2020ego}, CMU-Planner~\cite{cao2022aede}, and ART-Planner~\cite{wellhausen2023artplanner}.
EGO-Planner-2D denotes the planar implementation of EGO-Planner adapted for
ground-robot navigation. The evaluation covers three representative scenarios:
a random forest, a cluttered desk scene, and a stair scene. Each method is tested over
50 independent trials per scene, with maximum velocity \(1.5~\mathrm{m/s}\),
maximum acceleration \(1.0~\mathrm{m/s^2}\), and a timeout of \(120~\mathrm{s}\).
All simulations are run on an Intel Core i9-14900K CPU. The B-spline
optimization problems are solved using L-BFGS~\cite{liu1989limited}.
We report success rate (SR), collision rate (CR), and success weighted by path length (SPL).

\begin{table}[!t] \caption{Quantitative Comparison in Dense Obstacles} \label{tab:dense_obstacle_metrics} \centering \footnotesize \renewcommand{\arraystretch}{1.12} \setlength{\tabcolsep}{15.5pt} \begin{tabular}{@{}lccc@{}} \toprule \textbf{Method} & \textbf{SR}$\uparrow$ & \textbf{CR}$\downarrow$ & \textbf{SPL}$\uparrow$ \\ \midrule EGO-Planner-2D~\cite{zhou2020ego} & 0.96 & 0.04 & 0.88 \\ EGO-Planner-3D~\cite{zhou2020ego} & 0.76 & 0.24 & 0.75 \\ CMU-Planner~\cite{cao2022aede} & 0.92 & 0.08 & 0.86 \\ ART-Planner~\cite{wellhausen2023artplanner} & 0.44 & 0.56 & 0.31 \\ \midrule \textbf{SCAN-Planner (ours)} & \textbf{1.00} & \textbf{0.00} & \textbf{0.95} \\ \bottomrule \end{tabular} \end{table}

\subsection{Dense Obstacles}
\label{subsec:dense}
We first evaluate all planners in random dense-obstacle environments to assess their robustness against local minima and their ability to maintain smooth forward progress. Each map covers \(40~\mathrm{m}\times20~\mathrm{m}\), with the number of randomly placed obstacles increasing from 100 to 500. The obstacle width is sampled from \(0.2~\mathrm{m}\) to \(0.5~\mathrm{m}\), forming narrow passages with varying clearance margins for the quadruped body.

As shown in Fig.~\ref{fig:compare_2d}, all planners can find feasible paths in sparse scenes, while their efficiency and robustness diverge as the obstacle density increases. EGO-Planner-3D performs collision avoidance in the full 3D free space, which often introduces unnecessary vertical deformation for a ground robot. In contrast, EGO-Planner-2D plans on a planar representation and therefore produces more stable and shorter trajectories in this scenario. ART-Planner relies on random sampling, resulting in piecewise trajectories and frequent detours around distant obstacles. CMU-Planner adopts a sampling-based strategy with a predefined trajectory library. Although the selected trajectory is feasible, it is not necessarily optimal, leading to degraded SPL when narrow passages require precise route selection.

In terms of success rate and collision risk, EGO-Planner-2D, EGO-Planner-3D, and CMU-Planner approximate the robot body using a single inflation radius, which introduces an isotropic safety margin that does not capture the yaw-dependent anisotropy of the quadruped footprint. As a result, conservative inflation can unnecessarily eliminate traversable narrow passages, whereas insufficient inflation may leave residual collision risk in dense clutter. ART-Planner uses a rectangular robot model and thus provides more reliable collision checking, but its lack of trajectory optimization leads to increasingly longer paths as the environment becomes denser. In contrast, SCAN-Planner combines a yaw-aware twin-cylinder footprint with rebound-guided trajectory optimization. This design enables the planner to exploit yaw-dependent safe passages while avoiding excessive detours, maintaining smooth and efficient trajectories even when the number of obstacles increases to 500.

\subsection{3D Unstructured Scenes}
\label{subsec:unstructure}
\begin{figure}[!t]
    \centering
    \includegraphics[width=0.80\linewidth]{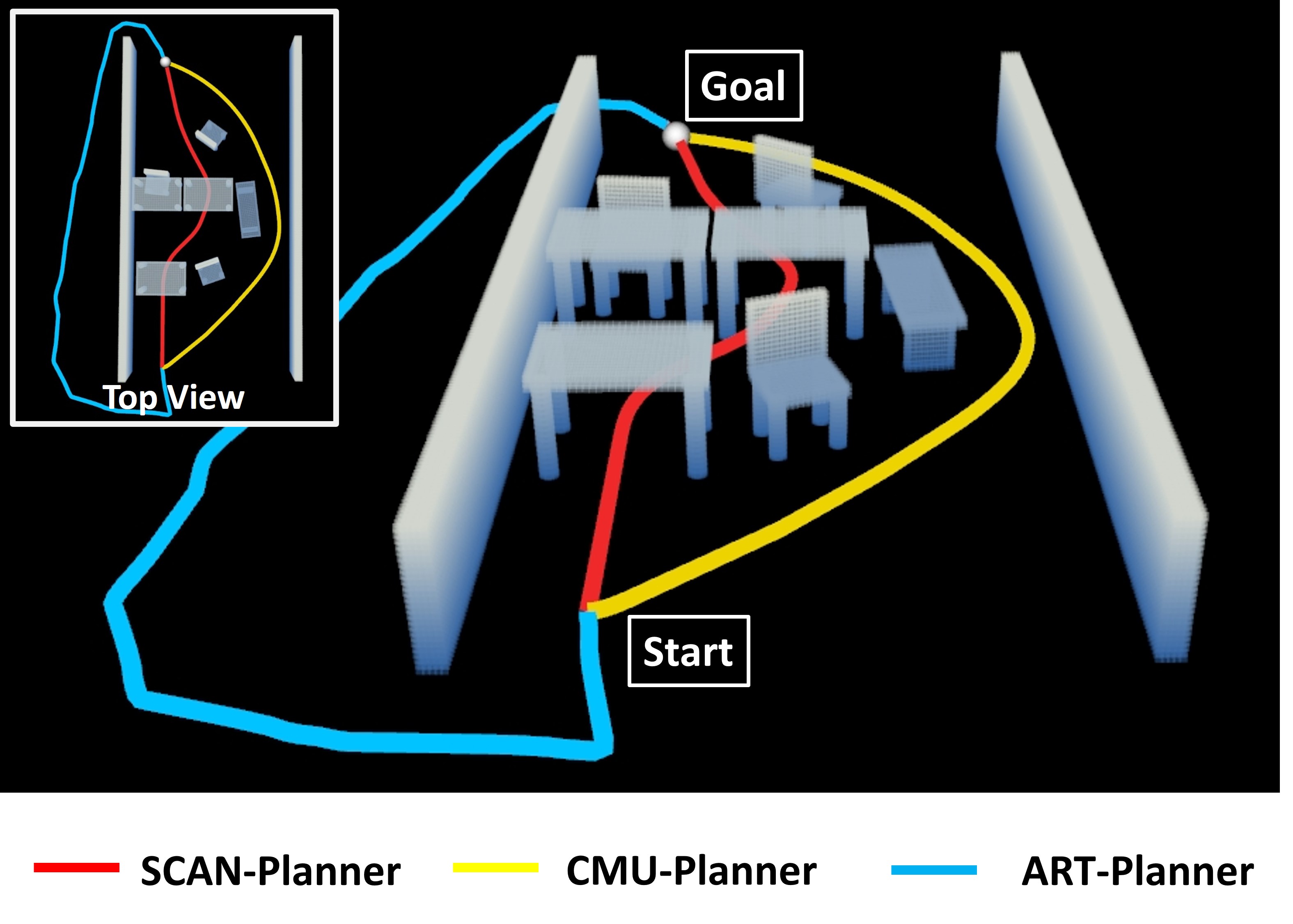}
    \caption{Comparison in a cluttered desk scene. SCAN-Planner (red) plans through the constrained 3D interior space by considering overhanging obstacles and local body clearance, while CMU-Planner (yellow) and ART-Planner (blue) produce larger detours around the structure.}
    \label{fig:compare_desk}
\end{figure}

Fig.~\ref{fig:compare_desk} shows a cluttered desk scene containing overhanging structures and constrained passages. This scene is challenging for planners relying on \(2.5\)D elevation maps, where tabletops are collapsed into a terrain-like representation and the under-table clearance cannot be explicitly constructed. As a result, CMU-Planner treats the desk region as occupied and selects a conservative bypass route. ART-Planner also relies on \(2.5\)D traversability reasoning and random sampling. Since a feasible passage through the constrained interior is not sampled early, the resulting solution detours around the outer wall, yielding the longest path. In contrast, SCAN-Planner maintains a full 3D occupancy representation and evaluates a yaw-aware 3D footprint of the quadruped body. It can explicitly check the body clearance under the overhanging desk and therefore selects the shortest route through the feasible interior passage.

\begin{figure}[!t]
    \centering
    \includegraphics[width=\linewidth]{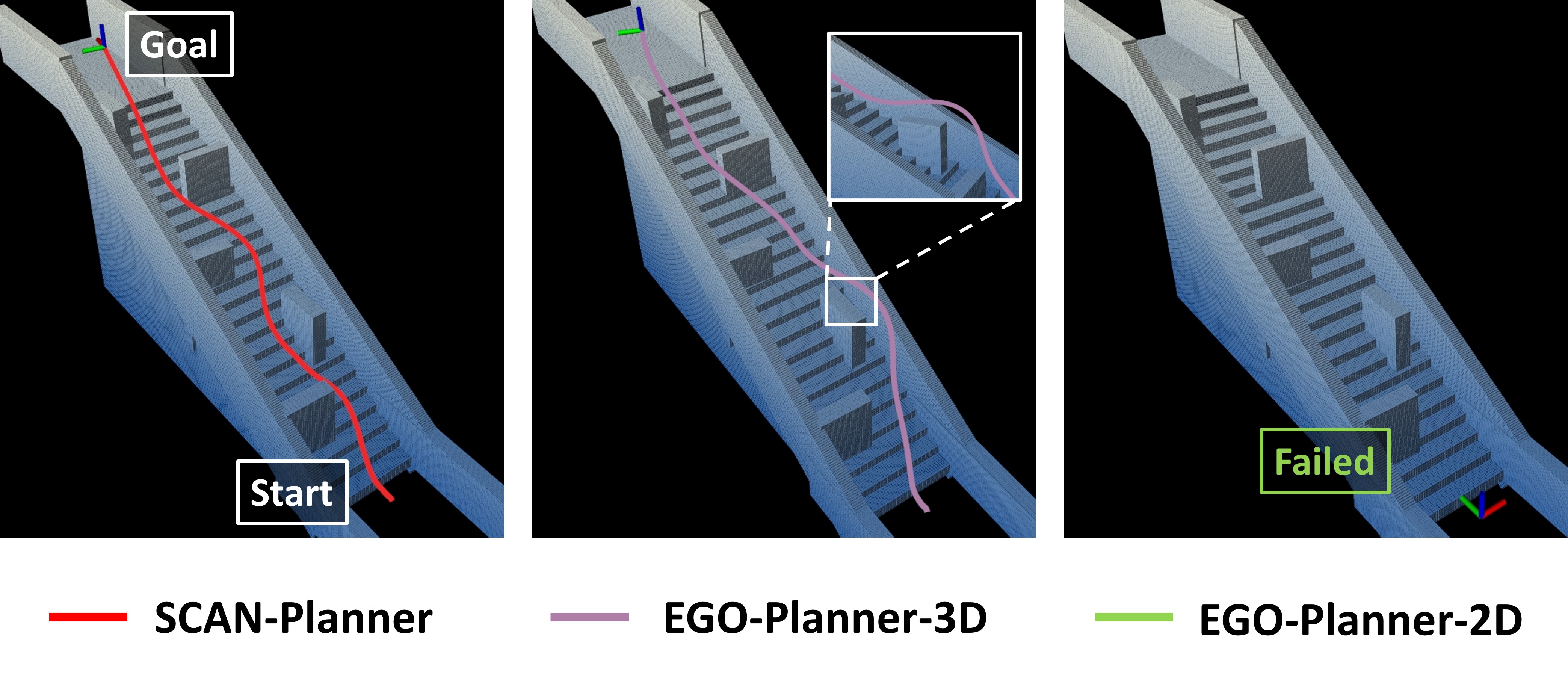}
    \caption{Comparison in a stair scene with obstacles. SCAN-Planner (red) maintains a stable stair-following trajectory while avoiding obstacles. EGO-Planner-3D (purple) avoids the obstacle with excessive vertical deformation, and EGO-Planner-2D (green) fails because the staircase is projected as an obstacle in its planar representation.}
    \label{fig:compare_stair}
\end{figure}

\begin{figure}[!t]
    \centering
    \includegraphics[width=\linewidth]{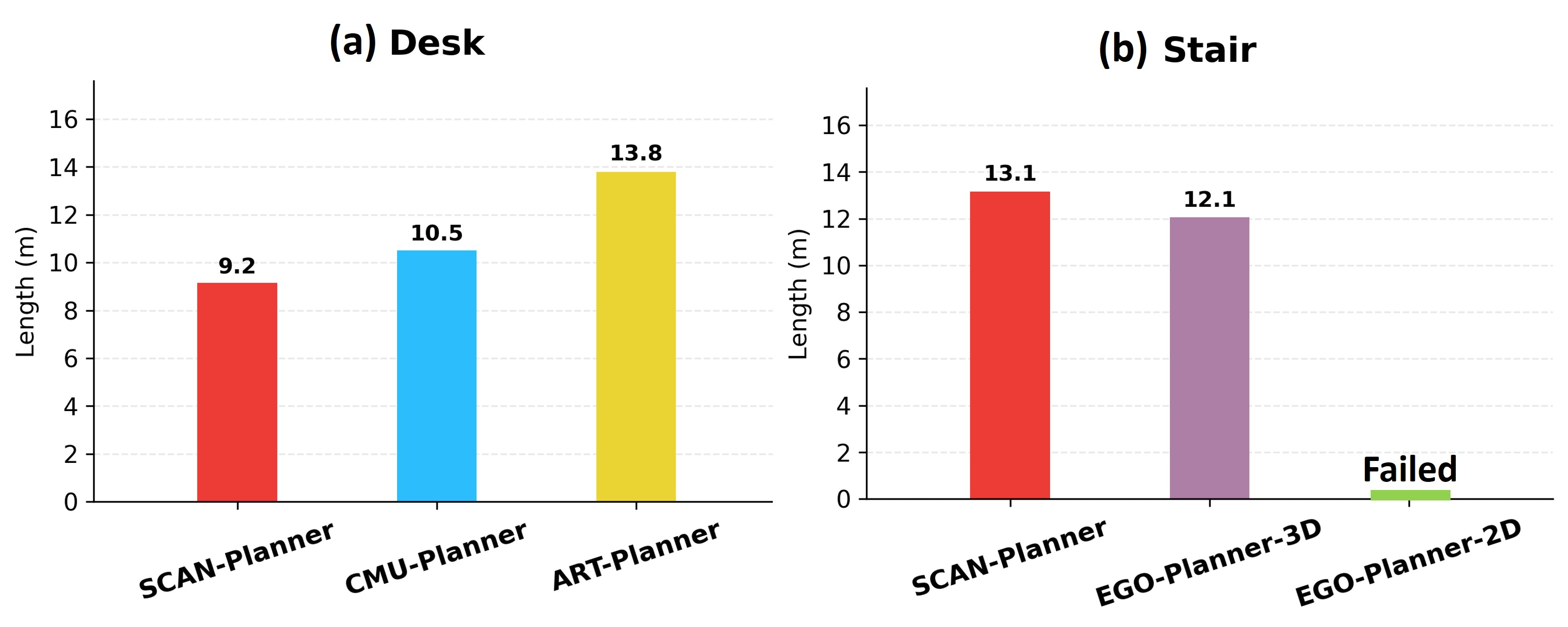}
    \caption{Path length comparison in representative 3D unstructured scenarios. (a) In the desk scene, SCAN-Planner achieves the shortest feasible path by exploiting the under-table clearance, whereas CMU-Planner and ART-Planner take longer bypass routes. (b) In the stair scene, EGO-Planner-3D yields a shorter path by passing over the obstacle, which is infeasible for contact-supported locomotion. EGO-Planner-2D fails because the staircase is treated as an obstacle in its 2D occupancy representation.} \label{fig:compare_length}
\end{figure}

The stair scenario in Fig.~\ref{fig:compare_stair} further evaluates vertical traversal and 3D obstacle avoidance. In this case, the planner must avoid the obstacle during stair traversal while preserving the nominal stair-following height profile. EGO-Planner-3D can reach the goal by planning in the full 3D free space. Although it produces a shorter path length in Fig.~\ref{fig:compare_length}, this reduction is achieved by deforming the trajectory upward and passing over the obstacle, which is infeasible for a ground robot constrained to contact-supported locomotion. When EGO-Planner-2D is used, it produces stable trajectories in planar clutter but cannot model stair traversability, causing the staircase to be treated as an obstacle. In contrast, SCAN-Planner preserves the vertical trend induced by the stairs and suppresses unnecessary rebound in the \(z\)-direction. The optimized trajectory therefore avoids the obstacle mainly through horizontal deformation while remaining compatible with ground locomotion.

\subsection{Real-World Experiments}
\label{subsec:real}
In the real-world experiments, we deploy SCAN-Planner on a Unitree Go2 quadruped robot, as shown in Fig.~\ref{fig:real_teaser}. The robot's dynamic limits are set to \(v_{\mathrm m}=1.0~\mathrm{m/s}\) and \(a_{\mathrm m}=0.25~\mathrm{m/s^2}\). To obtain dense and high-quality point clouds, the built-in LiDAR of the Go2 is replaced with a Livox Mid-360 LiDAR. An adapted FAST-LIO2~\cite{xu2022fast} is used for high-frequency state estimation. All perception, mapping, planning, and control modules run onboard an NVIDIA Jetson Orin NX in real time.

\begin{figure}[!t]
    \centering
    \includegraphics[width=\linewidth]{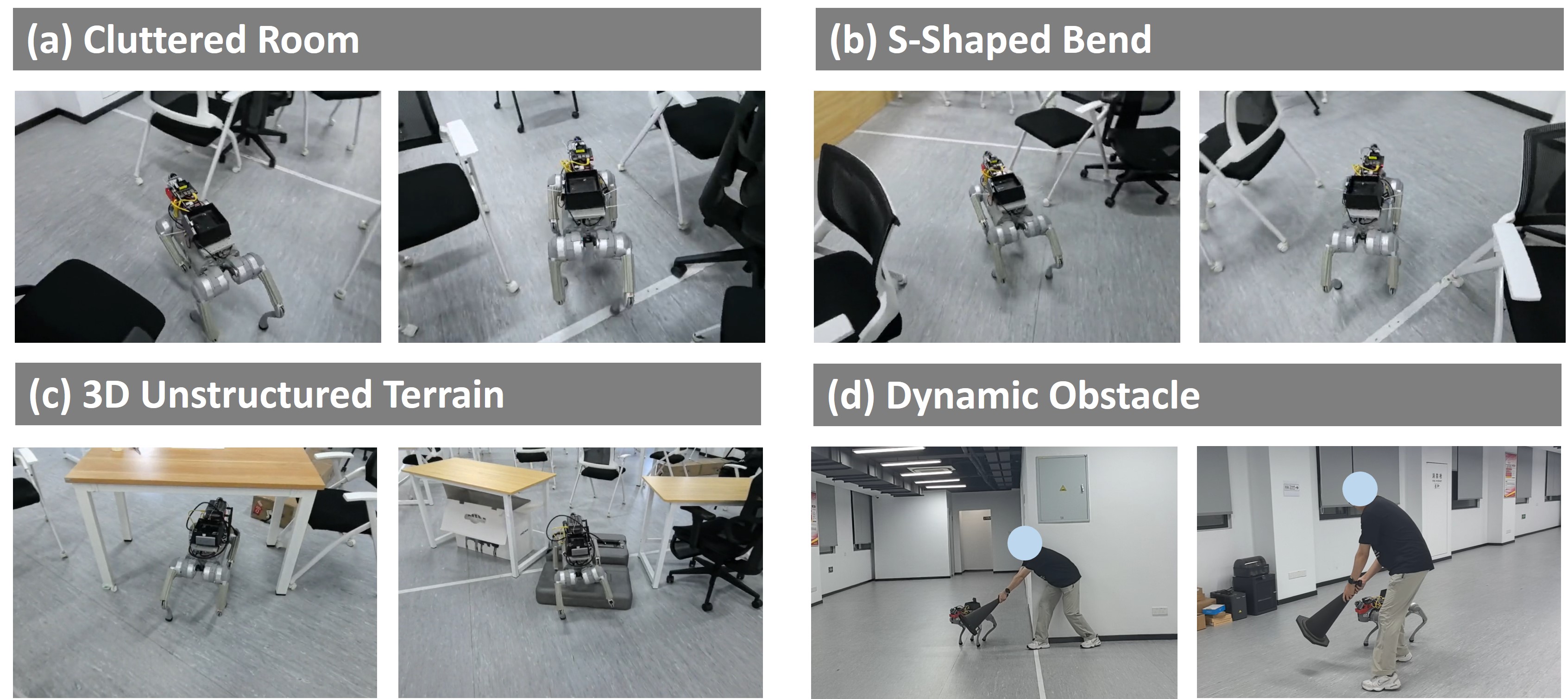}
    \caption{Indoor real-world experiments of SCAN-Planner in challenging environments. (a) Cluttered room navigation, where the quadruped robot safely passes through narrow passages formed by desks and chairs. (b) S-shaped bend navigation, where consecutive sharp turns and confined corridors evaluate the smoothness and robustness of the planned trajectory. (c) 3D unstructured terrain traversal, where the robot reasons about tables, chairs, and low obstacles to distinguish traversable and non-traversable regions. (d) Dynamic obstacle avoidance, where the planner updates the local occupancy map online and generates safe motions around pedestrians and other moving obstacles.}
    \label{fig:real_four}
\end{figure}

We first conduct four indoor experiments in challenging environments to evaluate the local navigation capability of the proposed planner. In the cluttered-room experiment shown in Fig.~\ref{fig:real_four}(a), the robot navigates from an entrance to an exit through multiple narrow passages formed by desks and chairs.
The results demonstrate that SCAN-Planner can generate safe and efficient collision-free motions in densely cluttered indoor scenes.
In the S-shaped bend experiment shown in Fig.~\ref{fig:real_four}(b), consecutive sharp turns and confined corridors are used to evaluate trajectory smoothness and robustness.
The planned trajectories remain smooth and continuous throughout the task, allowing the robot to traverse the maze-like passage without oscillation or deadlock.
We further evaluate the planner in a 3D unstructured environment, as shown in Fig.~\ref{fig:real_four}(c), where the scene contains traversable tables, regions blocked by chairs, and low obstacles that can be stepped over.
By combining the 3D occupancy representation with yaw-aware body collision checking, the robot can distinguish feasible clearances from blocked regions and select shorter, more efficient paths.
Finally, we test the system in dynamic environments with human disturbances, as shown in Fig.~\ref{fig:real_four}(d). The log-odds occupancy update enables the local map to adapt to environmental changes, allowing the robot to safely and smoothly avoid slowly moving obstacles during navigation.

To further validate the scalability of SCAN-Planner in long-range navigation tasks, we conduct two large-scale real-world experiments. The first experiment is a cross-floor inspection task in a three-story office building, where predefined inspection waypoints are placed at key locations on each floor. The task covers a volume of approximately $[45 \times 15 \times 15]~\mathrm{m}^3$, takes \(251~\mathrm{s}\) to complete, and results in a traveled distance of \(149~\mathrm{m}\). A representative point-cloud map and the corresponding 3D navigation trajectory are shown in Fig.~\ref{fig:real_teaser}. During the experiment, the robot reliably traverses staircases and avoids temporary obstacles while maintaining stable navigation performance. The second experiment is a last-mile delivery task in a campus-scale outdoor environment. The navigation region spans approximately $[278 \times 48 \times 5]~\mathrm{m}^3$, and the robot travels \(367~\mathrm{m}\) in \(589~\mathrm{s}\). A coarse global route is provided by a commercial navigation map, while SCAN-Planner performs local planning. During the task, the robot safely avoids pedestrians, bicycles, and large vehicles, and reaches the destination efficiently. These results show that the proposed sliding-map representation enables accurate local obstacle avoidance while remaining scalable to long-range navigation in large-scale indoor and outdoor environments. More detailed experimental results and visualizations are provided in the \textbf{supplementary video}.

\section{Conclusion}
\label{sec:conclusion}
This paper presented SCAN-Planner, a spatial collision-aware local planning framework for route-guided long-range quadruped navigation.
The proposed method addresses three practical requirements for deploying quadruped robots in cluttered and unstructured environments: body-aware motion in narrow spaces, stable 3D planning for ground robots, and scalable local planning under long-range route guidance.
To this end, SCAN-Planner introduces a yaw-aware twin-cylinder footprint for efficient whole-body collision checking, a projected A$^\ast$ rebound search with z-gradient suppression for horizontally dominated obstacle avoidance, and a robot-centric sliding map with dead-end recovery for high-resolution large-scale navigation.
Simulation and real-world experiments demonstrate that the planner generates smooth, safe, and efficient trajectories in dense obstacle fields, overhanging 3D scenes, stair environments, and long-range indoor and outdoor tasks.
Future work will further incorporate terrain analysis and autonomous exploration capabilities, enabling goal-free 3D scene navigation without predefined target points.

\bibliographystyle{IEEEtran}
\bibliography{IEEEabrv, example}  

@STRING{ICRA        = "ICRA."}

@STRING{IROS        = "IROS."}

@STRING{RAL        = "{IEEE} Robot. Autom. Lett."}

@STRING{IEEE_J_RO         = "{IEEE} Trans. Robot."}

@inproceedings{geng2023robo,
  title={Robo-Centric ESDF: A fast and accurate whole-body collision evaluation tool for any-shape robotic planning},
  author={Geng, Shuang and Wang, Qianhao and Xie, Lei and Xu, Chao and Cao, Yanjun and Gao, Fei},
  booktitle={IROS},
  pages={290--297},
  year={2023},
  organization={IEEE}
}

@inproceedings{frey2022locomotion,
  title={Locomotion Policy Guided Traversability Learning Using Volumetric Representations of Complex Environments},
  author={Frey, Jonas and Hoeller, David and Khattak, Shehryar and Hutter, Marco},
  booktitle={IROS},
  pages={5722--5729},
  year={2022},
  organization={IEEE},
  doi={10.1109/IROS47612.2022.9982190}
}

@article{kong2022marsim,
  title={{MARSIM}: A Light-Weight Point-Realistic Simulator for LiDAR-Based UAVs},
  author={Kong, Fanze and Liu, Xiyuan and Tang, Benxu and Lin, Jiarong and Ren, Yunfan and Cai, Yixi and Zhu, Fangcheng and Chen, Nan and Zhang, Fu},
  journal={arXiv preprint arXiv:2211.10716},
  year={2022}
}

@article{zhou2020ego,
  title={EGO-Planner: An ESDF-free Gradient-based Local Planner for Quadrotors},
  author={Zhou, Xin and Wang, Zhepei and Ye, Hongkai and Xu, Chao and Gao, Fei},
  journal=RAL,
  volume={6},
  number={2},
  pages={478--485},
  year={2021},
  publisher={IEEE},
  doi={10.1109/LRA.2020.3047728}
}

@article{zhou2019robust,
  title={Robust and Efficient Quadrotor Trajectory Generation for Fast Autonomous Flight},
  author={Zhou, Boyu and Gao, Fei and Wang, Luqi and Liu, Chuhao and Shen, Shaojie},
  journal=RAL,
  volume={4},
  number={4},
  pages={3529--3536},
  year={2019},
  publisher={IEEE},
  doi={10.1109/LRA.2019.2927938}
}

@article{tordesillas2022faster,
  title={{FASTER}: Fast and Safe Trajectory Planner for Navigation in Unknown Environments},
  author={Tordesillas, Jesus and Lopez, Brett T. and Everett, Michael and How, Jonathan P.},
  journal=IEEE_J_RO,
  volume={38},
  number={2},
  pages={922--938},
  year={2022},
  publisher={IEEE},
  doi={10.1109/TRO.2021.3100142}
}

@article{wang2022geometrically,
  title={Geometrically Constrained Trajectory Optimization for Multicopters},
  author={Wang, Zhepei and Zhou, Xin and Xu, Chao and Gao, Fei},
  journal=IEEE_J_RO,
  volume={38},
  number={5},
  pages={3259--3278},
  year={2022},
  publisher={IEEE},
  doi={10.1109/TRO.2022.3160022}
}

@article{han2021fast,
  title={Fast-Racing: An Open-Source Strong Baseline for {SE(3)} Planning in Autonomous Drone Racing},
  author={Han, Zhichao and Wang, Zhepei and Pan, Neng and Lin, Yi and Xu, Chao and Gao, Fei},
  journal=RAL,
  volume={6},
  number={4},
  pages={8631--8638},
  year={2021},
  publisher={IEEE},
  doi={10.1109/LRA.2021.3113976}
}

@article{zhou2021raptor,
  title={{RAPTOR}: Robust and Perception-Aware Trajectory Replanning for Quadrotor Fast Flight},
  author={Zhou, Boyu and Pan, Jie and Gao, Fei and Shen, Shaojie},
  journal=IEEE_J_RO,
  volume={37},
  number={6},
  pages={1992--2009},
  year={2021},
  publisher={IEEE},
  doi={10.1109/TRO.2021.3071527}
}

@article{liu1989limited,
  title={On the Limited Memory {BFGS} Method for Large Scale Optimization},
  author={Liu, Dong C. and Nocedal, Jorge},
  journal={Mathematical Programming},
  volume={45},
  number={1},
  pages={503--528},
  year={1989},
  publisher={Springer},
  doi={10.1007/BF01589116}
}

@article{chen2023smug,
  author  = {Changan Chen and Jonas Frey and Philip Arm and Marco Hutter},
  title   = {{SMUG} Planner: A Safe Multi-Goal Planner for Mobile Robots in Challenging Environments},
  journal = {IEEE Robotics and Automation Letters},
  volume  = {8},
  number  = {11},
  pages   = {7170--7177},
  year    = {2023},
  doi     = {10.1109/LRA.2023.3311207}
}

@inproceedings{miki2022elevation,
  author    = {Takahiro Miki and Lorenz Wellhausen and Ruben Grandia and Fabian Jenelten and Timon Homberger and Marco Hutter},
  title     = {Elevation Mapping for Locomotion and Navigation Using {GPU}},
  booktitle = {2022 IEEE/RSJ International Conference on Intelligent Robots and Systems (IROS)},
  pages     = {2273--2280},
  year      = {2022},
  doi       = {10.1109/IROS47612.2022.9981507}
}

@inproceedings{wellhausen2021rough,
  author    = {Lorenz Wellhausen and Marco Hutter},
  title     = {Rough Terrain Navigation for Legged Robots Using Reachability Planning and Template Learning},
  booktitle = {2021 IEEE/RSJ International Conference on Intelligent Robots and Systems (IROS)},
  pages     = {6914--6921},
  year      = {2021},
  doi       = {10.1109/IROS51168.2021.9636358}
}

@article{grandia2023perceptive,
  author  = {Ruben Grandia and Fabian Jenelten and Shaohui Yang and Farbod Farshidian and Marco Hutter},
  title   = {Perceptive Locomotion Through Nonlinear Model-Predictive Control},
  journal = {IEEE Transactions on Robotics},
  volume  = {39},
  number  = {5},
  pages   = {3402--3421},
  year    = {2023},
  doi     = {10.1109/TRO.2023.3275384}
}

@article{dong2025marg,
  author  = {Yinzhao Dong and Ji Ma and Liu Zhao and Wanyue Li and Peng Lu},
  title   = {{MARG}: {MA}stering Risky Gap Terrains for Legged Robots With Elevation Mapping},
  journal = {IEEE Transactions on Robotics},
  volume  = {41},
  pages   = {6123--6139},
  year    = {2025},
  doi     = {10.1109/TRO.2025.3619041}
}

@article{li2026stable,
  author  = {Congfei Li and Shuyue Lin and Shenwei Qu and Zhuoyuan Liu and Qingjun Yang and Max Q.-H. Meng and Yuxiang Sun},
  title   = {Stable Trajectory Planning for Quadruped Robots Using Terrain Features at Feet End},
  journal = {IEEE Robotics and Automation Letters},
  volume  = {11},
  number  = {2},
  pages   = {2266--2273},
  year    = {2026},
  doi     = {10.1109/LRA.2025.3645657}
}

@article{son2026roa,
  author  = {Yeongwoo Son and Hyunyong Lee and Hansol Kang and Jiman Park and SeongWon Nam and Jaeyoung Oh and Bumsu Yi and Hyeonwoo Yu and Hyouk Ryeol Choi},
  title   = {{3D RoA-Planner}: Path Planner for Quadruped Robots in Confined Spaces Using {3D} Rotatable Areas},
  journal = {IEEE Robotics and Automation Letters},
  volume  = {11},
  number  = {6},
  pages   = {7436--7443},
  year    = {2026},
  doi     = {10.1109/LRA.2026.3688059}
}

@inproceedings{ren2023rogmap,
  author  = {Yunfan Ren and Yixi Cai and Fangcheng Zhu and Siqi Liang and Fu Zhang},
  title   = {{ROG-Map}: An Efficient Robocentric Occupancy Grid Map for Large-scene and High-resolution LiDAR-based Motion Planning},
  booktitle = {2024 IEEE/RSJ International Conference on Intelligent Robots and Systems (IROS)},
  pages   = {8119--8125},
  year    = {2024},
  doi     = {10.1109/IROS58592.2024.10802303}
}

@inproceedings{zhang2025traversableplanes,
  author    = {Mengke Zhang and Zhihao Tian and Yaoguang Xia and Chao Xu and Fei Gao and Yanjun Cao},
  title     = {Efficient Trajectory Generation Based on Traversable Planes in 3{D} Complex Architectural Spaces},
  booktitle = {2025 IEEE International Conference on Robotics and Automation (ICRA)},
  pages     = {14513--14519},
  year      = {2025},
  doi       = {10.1109/ICRA55743.2025.11128727}
}

@article{li2025multilevel,
  author  = {Yuxiang Li and Kun Chen and Yifei Wang and Weifan Zhang and Jiancheng Wang and Haoyao Chen and Yunhui Liu},
  title   = {Real-Time Multilevel Terrain-Aware Path Planning for Ground Mobile Robots in Large-Scale Rough Terrains},
  journal = {IEEE Transactions on Robotics},
  volume  = {41},
  pages   = {4159--4179},
  year    = {2025},
  doi     = {10.1109/TRO.2025.3577015}
}

@article{zhang2024falcon,
  author  = {Yichen Zhang and Xinyi Chen and Ching Mei Feng and Boyu Zhou and Shaojie Shen},
  title   = {{FALCON}: Fast Autonomous Aerial Exploration Using Coverage Path Guidance},
  journal = {IEEE Transactions on Robotics},
  volume  = {41},
  pages   = {1365--1385},
  year    = {2025},
  doi     = {10.1109/TRO.2024.3522148}
}

@article{wellhausen2023artplanner,
  author  = {Lorenz Wellhausen and Marco Hutter},
  title   = {{ArtPlanner}: Robust Legged Robot Navigation in the Field},
  journal = {Field Robotics},
  volume  = {3},
  pages   = {413--434},
  year    = {2023},
  doi     = {10.55417/fr.2023013}
}

@inproceedings{cao2022aede,
  author    = {Chao Cao and Hongbiao Zhu and Fan Yang and Yukun Xia and Howie Choset and Jean Oh and Ji Zhang},
  title     = {Autonomous Exploration Development Environment and the Planning Algorithms},
  booktitle = {2022 IEEE International Conference on Robotics and Automation (ICRA)},
  pages     = {8921--8928},
  year      = {2022},
  doi       = {10.1109/ICRA46639.2022.9812330}
}

@article{xu2022fast,
  title={Fast-lio2: Fast direct lidar-inertial odometry},
  author={Xu, Wei and Cai, Yixi and He, Dongjiao and Lin, Jiarong and Zhang, Fu},
  journal={IEEE Transactions on Robotics},
  volume={38},
  number={4},
  pages={2053--2073},
  year={2022},
  publisher={IEEE},
  doi={10.1109/TRO.2022.3141876}
}

@article{tang2026memory,
  title={Memory-efficient boundary map for large-scale occupancy grid mapping},
  author={Tang, Benxu and Ren, Yunfan and Cai, Yixi and Kong, Fanze and Liu, Wenyi and Zhu, Fangcheng and Yin, Longji and Shi, Liuyu and Zhang, Fu},
  journal={The International Journal of Robotics Research},
  pages={02783649261425266},
  year={2026},
  publisher={SAGE Publications Sage UK: London, England}
}

\end{document}